\documentclass{article}

\usepackage{iclr2027_conference,times}

\usepackage[utf8]{inputenc}
\usepackage[T1]{fontenc}
\usepackage{microtype}
\usepackage{amsmath,amssymb,mathtools}
\usepackage{amsthm}
\usepackage{booktabs}
\usepackage{array}
\usepackage{graphicx}
\usepackage{xcolor}
\usepackage{algorithm}
\usepackage{algpseudocode}
\usepackage[colorlinks=true,citecolor=blue,linkcolor=black,urlcolor=black]{hyperref}
\usepackage{url}

\newcommand{\Var}{\operatorname{Var}}
\newcommand{\Bern}{\operatorname{Bernoulli}}

\title{Speculative Evaluation of Stochastic LLMs}

\author{Qianli Shen\textsuperscript{1}, Xiang Li\textsuperscript{2}\thanks{\fontsize{8}{9.5}\selectfont Work done during internships at Alibaba Group\quad\textsuperscript{$\dagger$}Correspondence to: \{daoyuanchen.cdy, yaliang.li\}@alibaba-inc.com\\\hspace*{11.88pt}Our code is released at {\urlstyle{same}\url{https://github.com/ShenQianli/SpecEval}}.}\,, Ruomeng Ding\textsuperscript{3}\footnotemark[1]\,,
Yanxi Chen\textsuperscript{1}, Daoyuan Chen\textsuperscript{1}\footnotemark[2]\,, Yaliang Li\textsuperscript{1}\footnotemark[2]\\
\multicolumn{1}{l}{\small\textsuperscript{1}Alibaba Group\quad
\textsuperscript{2}National University of Singapore\quad
\textsuperscript{3}University of North Carolina at Chapel Hill}}

\iclrfinalcopy

\begin{document}

\maketitle
\fancyhead{}
\vspace{-10pt}

\begin{abstract}
Evaluating a stochastic large language model is costly: benchmark scores
estimate expected performance from randomized rollouts, yet uniform repetition
ignores sharp differences in task-level rollout variance.  We ask how to
minimize the variance of a fixed-benchmark mean under an exact rollout budget.
We develop Speculative Evaluation with a Hierarchical Bayesian Neyman (HBN)
policy with pilot size and stage weight jointly chosen ex ante.  It runs a short
uniform pilot, pools
per-task success counts with a hierarchical Bayesian model, and uses posterior
expectations of task-level sampling variances for
exact positive-integer Neyman allocation.
To mitigate the pilot synchronization barrier, HBN-async speculatively
executes continuations from partial pilot feedback and retains those selected
by the final allocation.  Across
six checkpoints and 18 benchmark groups, we evaluate 107 nondegenerate
benchmark--checkpoint profiles.  For rollout budgets of 8--64 per task,
Speculative Evaluation reduces variance relative to Uniform by
12.8\%--33.6\% on average across profiles, outperforming hindsight-tuned
empirical and independent Bayesian baselines.  Real-generation experiments that account for the
pilot synchronization barrier show that HBN-async mitigates its overhead,
helping translate statistical efficiency into practical evaluation benefits.
\end{abstract}

\section{Introduction}
Benchmarking a stochastic large language model (LLM) is a statistical
estimation problem.  Rollouts on the same task can produce different
answers, so a benchmark score estimates expected performance rather than a
deterministic quantity.  Quantifying this uncertainty requires repeated
generation, which is often omitted because of its cost
\citep{blackwell2024reproducible}.  Even with substantial repetition,
uncertainty can remain appreciable.  For the AIME24--Qwen3.5-4B profile studied
in Section~\ref{sec:experiments}, uniform evaluation with \(b=32\) rollouts per
task---a common choice for AIME---still yields a
normal-reference 95\% confidence interval with a half-width of \(1.85\)
benchmark percentage points.
Under uniform evaluation with \(b\) rollouts per task, the standard error
decreases only as \(b^{-1/2}\); halving an error bar therefore requires roughly
four times as many rollouts.  Reliable stochastic evaluation can therefore
require large rollout budgets.

Automated research and self-improvement make both evaluation cost and
precision especially important.  Systems such as AFlow \citep{zhang2025aflow},
SICA \citep{robeyns2025sica}, and the Darwin G\"odel Machine
\citep{zhang2026darwin} repeatedly propose and evaluate changes to LLM
workflows or agents, making evaluation a recurring cost rather than a
one-time measurement.  Their candidate changes may yield only small
improvements, so evaluation must also provide sufficiently precise feedback
to distinguish genuine progress from rollout noise.  Otherwise, search may
retain ineffective changes or discard useful ones.  Increasing repetition
improves precision but raises the cost of every iteration, motivating more
informative evaluation under a fixed rollout budget.

This motivates a more basic question that applies beyond auto-research:
\emph{can a fixed rollout budget be allocated more efficiently?}  Uniform
evaluation gives every task the same number of samples, although task-level
rollout variances can differ sharply.  For a binary outcome with pass
probability \(p_i\), the variance is \(p_i(1-p_i)\): tasks near \(p_i=1/2\)
are intrinsically noisy, while tasks near zero or one require fewer samples to
estimate precisely.  If known, these variances would let classical Neyman
allocation \citep{neyman1934representative} concentrate rollouts on noisier tasks and minimize benchmark-mean
variance.  In practice they are unknown, and learning them consumes the same
budget that the allocation is meant to save.  A useful policy must therefore
balance the budget spent learning task variances against the gains from
adapting the allocation.

Following feasible two-stage Neyman designs
\citep{sukhatme1975allocation,cai2024performance}, we develop \emph{Speculative
Evaluation} for stochastic LLM benchmarks, using a \emph{Hierarchical Bayesian
Neyman (HBN)} policy.  A uniform pilot observes every
task, and a hierarchical Bayesian model pools its noisy counts to construct posterior
expected sampling-variance scores.  An exact integer Neyman allocator
distributes the remaining budget while retaining at least one fresh rollout
per task.  Pilot size and stage weight are jointly chosen ex ante; fresh
continuation samples permit unbiased reuse of the pilot through a weighted
estimator.

Statistical efficiency alone does not ensure runtime efficiency.  Finalizing
the allocation requires all pilot rewards, creating a synchronization barrier
that can leave compute idle.  We therefore distinguish \emph{HBN-sync},
which waits before dispatching continuations, from \emph{HBN-async}, which
uses partial pilot feedback to speculatively pre-allocate and execute them.
The final allocation determines which requests are retained.  Although
unselected speculation wastes computation, executing continuations early
can reduce idle time at the barrier, improving throughput and overall
evaluation efficiency.

Speculation thus operates at two levels.  At the statistical level,
speculation uses pilot evidence to predict which tasks will benefit from
more samples.  At the systems level, speculation uses partial pilot feedback
to predict which continuations will be retained and executes them early.
The first can reduce estimator variance, while the second
helps translate statistical efficiency into wall-clock efficiency without
changing the estimator.

Our contributions are threefold:
\begingroup
\setlength{\leftmargini}{1.8em}
\setlength{\itemsep}{1pt}
\setlength{\parsep}{0pt}
\setlength{\topsep}{2pt}
\setlength{\partopsep}{0pt}
\begin{itemize}
  \item We formulate stochastic LLM evaluation as variance-minimizing integer
  allocation.  We develop an unbiased two-stage HBN policy with ex-ante pilot and
  weight design, hierarchical information sharing, and exact integer allocation.
  \item We introduce HBN-async, which uses partial pilot feedback to
  speculatively execute continuations, mitigating the synchronization barrier
  so that statistical gains can translate into practical runtime benefits.
  \item For rollout budgets of 8--64 per task across 107 benchmark--checkpoint
  profiles, HBN reduces variance by
  \(12.8\%\)--\(33.6\%\) on average relative to Uniform, outperforming tuned
  empirical and independent Bayesian baselines on this metric.  Real-generation
  experiments show that HBN-async mitigates barrier overhead, helping turn
  statistical efficiency into practical gains.
\end{itemize}
\endgroup

\section{Background}

\begingroup
\setlength{\parskip}{2pt}
\setlength{\abovedisplayskip}{2pt plus 1pt minus 1pt}
\setlength{\belowdisplayskip}{2pt plus 1pt minus 1pt}
\setlength{\abovedisplayshortskip}{0pt}
\setlength{\belowdisplayshortskip}{1pt}
\paragraph{Problem formulation.}
\label{sec:problem}

Fix a checkpoint and a benchmark of \(N\) tasks.  Rollout \(t\) on task \(i\)
produces \(Y_{i,t}\sim\Bern(p_i)\), for \(i=1,\ldots,N\),
independently conditional on the pass-probability vector
\(\mathbf p=(p_1,\ldots,p_N)\).  Our target is the benchmark mean
\(\mu(\mathbf p)=N^{-1}\sum_{i=1}^N p_i\).  We call \(\mathbf p\) the
task-probability profile of this benchmark--checkpoint pair.

Let \(b\in\mathbb Z_{\geq1}\) be the average rollout budget per task.
An allocation \(\mathbf n\in\mathbb Z_{\geq1}^N\) satisfies
\(\sum_{i=1}^N n_i=Nb\).  Under this allocation, define
\(\widehat p_i=n_i^{-1}\sum_{t=1}^{n_i}Y_{i,t}\) and
\(\widehat\mu(\mathbf n)=N^{-1}\sum_{i=1}^N\widehat p_i\).
The estimator is conditionally unbiased, with
\begin{equation}
  \Var\!\left(\widehat\mu\mid\mathbf p,\mathbf n\right)
  =\frac{1}{N^2}\sum_{i=1}^N\frac{\sigma_i^2}{n_i},
  \qquad \sigma_i^2=p_i(1-p_i).
  \label{eq:conditional-variance}
\end{equation}
Hence the oracle allocation solves
\begin{equation}
  \min_{\mathbf n\in\mathbb Z_{\geq1}^N}
  \sum_{i=1}^N\frac{\sigma_i^2}{n_i}
  \quad\text{subject to}\quad
  \sum_{i=1}^N n_i=Nb.
  \label{eq:allocation-problem}
\end{equation}
\endgroup
\vspace{-6pt}

\begingroup
\setlength{\parskip}{2pt}
\setlength{\abovedisplayskip}{2pt plus 1pt minus 1pt}
\setlength{\belowdisplayskip}{2pt plus 1pt minus 1pt}
\setlength{\abovedisplayshortskip}{0pt}
\setlength{\belowdisplayshortskip}{1pt}
\paragraph{Oracle Neyman allocation.}
\label{sec:oracle}

Uniform allocation sets \(n_i=b\) and has variance
\(V_{\mathrm{unif}}(\mathbf p)=(N^2b)^{-1}\sum_i\sigma_i^2\).
For an analytic benchmark, relax integrality and the per-task lower bound
in \eqref{eq:allocation-problem}.  If an oracle knows \(\mathbf p\), the
classical Neyman allocation \citep{neyman1934representative} is
\(n_i^*=Nb\sigma_i/\sum_j\sigma_j\).
For profiles with positive total variance, its variance relative to Uniform is
\begin{equation}
  R_N:=\frac{V_{\mathrm{oracle}}}{V_{\mathrm{unif}}}
  =\frac{(\sum_i\sigma_i)^2}{N\sum_i\sigma_i^2}\leq1.
  \label{eq:oracle-ratio}
\end{equation}
\endgroup
The inequality follows from Cauchy--Schwarz.
Thus task heterogeneity creates an opportunity to reduce variance, but the
required \(\sigma_i\)'s are unknown.  A practical policy must first learn them
from pilot observations and then allocate the remaining budget.

\paragraph{Two-stage evaluation.}
\label{sec:two-stage}

Following classical feasible Neyman designs based on preliminary samples
\citep{sukhatme1975allocation,cai2024performance}, our two-stage policy requires
\(b\geq2\) and spends an integer pilot budget \(1\leq m\leq b-1\) on every task.  Let
\(S_i=\sum_{t=1}^m Y^{(0)}_{i,t}\)
be the pilot successes on task \(i\), and write
\(S=(S_1,\ldots,S_N)\) for the complete pilot vector.  The pilot determines a
positive integer
second-stage allocation \(\mathbf L(S)\) satisfying
\(\sum_{i=1}^N L_i(S)=N(b-m)\).
Thus the allocation may adapt to the pilot, while the second-stage rollouts
\(Y^{(1)}_{i,t}\) are fresh and conditionally independent of it.

Define the stage-wise estimators
\(\widehat\mu_0=(Nm)^{-1}\sum_i S_i\) and
\(\widehat\mu_1=N^{-1}\sum_i L_i(S)^{-1}
\sum_{t=1}^{L_i(S)}Y^{(1)}_{i,t}\).
The pilot can be recycled without bias using any deterministic stage weight
\(w\in[0,1]\) fixed before the target evaluation:
\begin{equation}
  \widehat\mu_w
  =w\widehat\mu_0+(1-w)\widehat\mu_1.
  \label{eq:two-stage-estimator}
\end{equation}
Since the pilot estimator is unbiased and the fresh continuation estimator
is unbiased conditional on the pilot, any stage weight \(w\) fixed before
evaluation yields an unbiased combined estimate.

Within this two-stage policy class, we seek precommitted \(m\), \(w\), and
an allocation rule \(\mathbf L(\cdot)\) whose realized allocation depends
only on the pilot:
\begin{equation*}
  \min_{\substack{m\in\mathbb Z,\ 1\leq m\leq b-1\\
                  0\leq w\leq1\\
                  \mathbf L(\cdot)\ \text{pilot-measurable}}}
  \Var(\widehat\mu_w\mid\mathbf p)
  \quad\text{subject to}\quad
  \mathbf L(S)\in\mathbb Z_{\geq1}^N,
  \quad \sum_{i=1}^N L_i(S)=N(b-m).
  \tag{P}
  \label{eq:two-stage-objective}
\end{equation*}
This design introduces one synchronization point before the final
continuation allocation.  More stages allow finer adaptation but add
coordination overhead; Section~\ref{sec:partial-allocation-prefetch} addresses
execution around the two-stage barrier.
Appendix~\ref{app:related-work} provides a comprehensive discussion of related
work, including how this choice relates to classical two-stage designs and
recent online allocation methods.

\section{Speculative Evaluation with Asynchronous Hierarchical Bayesian Neyman Allocation}
\label{sec:bayesian-neyman}

Because the task-probability profile in
Problem~\eqref{eq:two-stage-objective} is unknown, we choose the precommitted
pilot size and stage weight by minimizing prior-averaged risk rather than
profile-specific conditional variance.  A hierarchical Bayesian model yields
the Hierarchical Bayesian Neyman (HBN) policy.

\subsection{Statistical speculation: Hierarchical Bayesian Neyman allocation}
\label{sec:statistical-speculation}

\paragraph{Hierarchical variance estimation.}

Let \(\xi\in(0,1)\) represent the latent mean task success probability for the
benchmark--checkpoint pair, and let \(\delta\in(0,1)\) represent normalized
between-task dispersion.  We place independent hyperpriors \(\pi_\xi\) and
\(\pi_\delta\) on them, set \(\kappa=(1-\delta)/\delta\), and use
\begin{equation}
  p_i\mid\xi,\delta\overset{\mathrm{iid}}{\sim}
  \operatorname{Beta}\!\left(\xi\kappa,(1-\xi)\kappa\right).
  \label{eq:hierarchical-prior}
\end{equation}
Together, the hyperpriors and conditional task model induce a joint
hierarchical prior \(\pi\) on \((\xi,\delta,\mathbf p)\).
This parameterization gives \(\mathbb E[p_i\mid\xi,\delta]=\xi\),
\(\Var(p_i\mid\xi,\delta)=\xi(1-\xi)\delta\), and expected sampling variance
\(\mathbb E[p_i(1-p_i)\mid\xi,\delta]=\xi(1-\xi)(1-\delta)\).

The derivations allow any prespecified hyperprior, which affects allocation
efficiency but not unbiasedness.  All reported HBN analyses and experiments
use the untuned reference specification
\(\xi,\delta\overset{\mathrm{ind}}{\sim}\operatorname{Beta}(1,1)\).
Only the pilot updates the hyperparameter posterior.

For the pilot counts \(S=(S_1,\ldots,S_N)\) from Section~\ref{sec:two-stage},
let \(\mathrm B(a,b)\) denote the beta function.  Beta--binomial conjugacy
yields the hyperposterior
\begin{equation}
  \pi(\xi,\delta\mid S)
  \propto
  \pi_\xi(\xi)\pi_\delta(\delta)
  \prod_{i=1}^N
  \frac{\mathrm B(\xi\kappa+S_i,(1-\xi)\kappa+m-S_i)}
       {\mathrm B(\xi\kappa,(1-\xi)\kappa)}.
  \label{eq:hyperposterior}
\end{equation}
Neyman allocation depends on the conditional sampling variance
\(p_i(1-p_i)\) of a fresh rollout.  Since \(p_i\) is unknown, we use its
posterior expectation:
\begin{equation}
  \widetilde v_i(S)
  :=\mathbb E[p_i(1-p_i)\mid S]
  =\mathbb E_{\xi,\delta\mid S}\!\left[
    \frac{(\xi\kappa+S_i)((1-\xi)\kappa+m-S_i)}
         {(\kappa+m)(\kappa+m+1)}
  \right].
  \label{eq:posterior-rollout-variance}
\end{equation}
The full vector \(S\) updates the shared hyperparameters, while \(S_i\)
provides task-specific evidence.  Consequently, \(\widetilde v_i(S)\) remains
positive even when a task has only pilot successes or only pilot failures.

\paragraph{Posterior Neyman allocation.}
\label{sec:posterior-neyman-allocation}

After observing the complete pilot, minimizing posterior expected variance
gives \(L_i\propto\sqrt{\widetilde v_i(S)}\) when integrality and the
per-task lower bound are relaxed.  For actual evaluation, we solve
\begin{equation}
  \mathbf L^{\mathrm B}(S)
  \in\arg\min_{\substack{L_i\in\mathbb Z_{\geq1}\\
                         \sum_iL_i=N(b-m)}}
  \sum_{i=1}^N\frac{\widetilde v_i(S)}{L_i}.
  \label{eq:integer-bayesian-neyman}
\end{equation}
Starting from \(L_i=1\), assign each remaining rollout to the task with the
largest marginal reduction
\(\Delta_i(L_i)=\widetilde v_i(S)/[L_i(L_i+1)]\).
These gains decrease with \(L_i\), so the greedy procedure is the exact
integer optimum.  It preserves the budget and guarantees a fresh rollout for
every task.

\paragraph{Joint pilot-and-weight selection.}

For fixed \(\mathbf p\), the pilot and continuation estimators are uncorrelated:
the conditional mean of \(\widehat\mu_1\) given the pilot is the nonrandom
quantity \(\mu(\mathbf p)\).  Therefore any fixed \((m,w)\) and continuation
rule \(\mathbf L\) have exact conditional variance
\begin{equation}
\begin{gathered}
  \Var(\widehat\mu_w\mid\mathbf p)
  =w^2 A_m(\mathbf p)+(1-w)^2 B_m(\mathbf p;\mathbf L),\\
  A_m(\mathbf p)=\frac{1}{N^2m}\sum_i p_i(1-p_i),\qquad
  B_m(\mathbf p;\mathbf L)=\frac{1}{N^2}
  \mathbb E_{S\mid\mathbf p}\!\left[
    \sum_i\frac{p_i(1-p_i)}{L_i(S)}
  \right].
\end{gathered}
  \label{eq:two-stage-conditional-variance}
\end{equation}
HBN selects \((m,w)\) ex ante using only \((b,N)\) and the reference prior
\(\pi\).  Let \(\overline v_\pi=\mathbb E_\pi[p_i(1-p_i)]\).
For each candidate pilot, define the prior-averaged variance masses
without the common factor \(N^{-2}\):
\begin{equation}
  A_{\pi,m}=\frac{N\overline v_\pi}{m},
  \qquad
  B_{\pi,m}=\mathbb E_{S\sim\pi}\!\left[
    \sum_{i=1}^N\frac{\widetilde v_i(S)}{L_i^{\mathrm B}(S)}
  \right].
  \label{eq:prior-variance-masses}
\end{equation}
The expectation covers the hyperparameters, task probabilities, and pilot
outcomes.  Relative to the prior-averaged Uniform risk, the Bayes risk is
\begin{equation}
  \mathcal R_{\pi,N}(b,m,w)
  =\frac{b}{N\overline v_\pi}
   \left\{w^2A_{\pi,m}+(1-w)^2B_{\pi,m}\right\}.
  \label{eq:prior-predictive-risk}
\end{equation}
For each integer \(m\), the Bayes-optimal weight and minimized risk are
available in closed form:
\begin{equation}
  w_{\pi,m}^*=\frac{B_{\pi,m}}{A_{\pi,m}+B_{\pi,m}},
  \qquad
  \mathcal R_{\pi,N}^*(b,m)
  =\frac{b}{N\overline v_\pi}
   \frac{A_{\pi,m}B_{\pi,m}}{A_{\pi,m}+B_{\pi,m}}.
  \label{eq:bayes-optimal-stage-weight}
\end{equation}
We select \(m_{\mathrm B}^*\) by minimizing \(\mathcal R_{\pi,N}^*(b,m)\)
over integers \(m=1,\ldots,b-1\), then set
\(w_{\mathrm B}^*=w_{\pi,m_{\mathrm B}^*}^*\).
Both are fixed offline under the reference prior without target outcomes.
Algorithm~\ref{alg:bayesian-neyman} in Appendix~\ref{app:complete-algorithm}
summarizes the statistical policy.

\subsection{Systems speculation: Asynchronous pre-allocation}
\label{sec:partial-allocation-prefetch}

HBN introduces one statistical synchronization point: the continuation
allocation is finalized only after every pilot reward is available.  We
distinguish two execution schemes for the same statistical procedure.
\emph{HBN-sync} executes all pilot requests, waits for their rewards, computes
the final allocation, and only then dispatches continuations.  This execution
barrier can leave workers idle while waiting for the long-running pilot tail.
\emph{HBN-async} mitigates this barrier by using partial pilot feedback to
speculatively execute continuations early, without changing the statistical
design or final allocation rule.

At a pre-barrier time \(t\), task \(i\) has
\(n_i(t)\leq m\) scored pilot rollouts and \(S_i(t)\) successes, with at least
one \(n_i(t)<m\).  Write \(S(t)\) and \(\mathbf n(t)\) for the corresponding
vectors.  Replacing the common count \(m\) in
Eq.~\eqref{eq:hyperposterior} by the task-specific counts \(n_i(t)\) gives a
provisional posterior for pre-allocation.  Applying
Eq.~\eqref{eq:posterior-rollout-variance} under this distribution gives the
partial allocation scores:
\begin{equation}
  \widetilde v_i^{(t)}
  =\mathbb E_{\xi,\delta\mid S(t),\mathbf n(t)}\!\left[
    \frac{(\xi\kappa+S_i(t))
          ((1-\xi)\kappa+n_i(t)-S_i(t))}
         {(\kappa+n_i(t))(\kappa+n_i(t)+1)}
  \right].
  \label{eq:partial-posterior-rollout-variance}
\end{equation}
Substituting \(\widetilde v_i^{(t)}\) into
Eq.~\eqref{eq:integer-bayesian-neyman} gives a provisional allocation
\(\mathbf L^{\mathrm B,(t)}\), updated as pilot rewards arrive.  This
scheduling signal can be cost-sensitive if generation latency depends on
the outcome.  Once all designated pilot rewards are available, it recovers
the same final allocation rule as HBN-sync, without selection by completion
time.

The partial quantities above provide an online signal for managing HBN-async's
request queue.  We apply the following rules in priority order:
\begingroup
\setlength{\leftmargini}{\dimexpr\leftmargini/2\relax}
\begin{enumerate}
  \item \textbf{Pilot first.}  Every pilot request is submitted before any
  continuation request; pending pilot admission always takes precedence.
  \item \textbf{Allocation-gated continuations.}  For the \(\ell\)-th
  continuation of task \(i\), let
  \(r_{i,\ell}^{(t)}=\ell/L_i^{\mathrm B,(t)}\).  Only requests with
  \(r_{i,\ell}^{(t)}\leq1\) are queued, in increasing \(r_{i,\ell}^{(t)}\)
  order with deterministic tie breaking.  Partial updates refresh the pending
  queue without cancelling work already in flight.
  \item \textbf{Fixed-ID recovery.}  Pilot and continuation results are
  recovered by request IDs fixed before execution.  Even if another request
  for the same task finishes earlier, it cannot replace a designated ID,
  because such substitution can introduce latency-dependent selection bias.
\end{enumerate}
\endgroup
Once all pilot rewards are ready, the partial construction reduces to the
official \(\mathbf L^{\mathrm B}(S)\).  For each task, the router reuses the
first \(L_i^{\mathrm B}(S)\) continuation requests, submits any missing
requests in that prefix, and discards completed surplus work, cancels queued
surplus work, or aborts running surplus work.  Under the
conditional-independence assumption in Section~\ref{sec:two-stage}, unbiasedness
therefore remains unchanged.
Algorithm~\ref{alg:hbn-async} in Appendix~\ref{app:complete-algorithm}
summarizes the asynchronous execution scheme.

\section{Experiments}
\label{sec:experiments}

Our experiments separate statistical efficiency from execution cost.
Section~\ref{sec:variance-reduction} evaluates estimator variance conditional
on fixed empirical task-probability profiles.
Section~\ref{sec:system-overhead} measures generation and coordination costs
using real inference, with coupled rewards that give HBN-sync and HBN-async
identical final allocations.  We then combine the replay variance ratios
with measured runtimes in a variance-equivalent cost comparison.

\subsection{Benchmarks and rollout data}

\textbf{Models and benchmarks.}
We form the complete grid of six checkpoints and 18 benchmark groups, giving
108 benchmark--checkpoint pairs.  The checkpoints are Qwen2.5-3B, 7B, and
32B \citep{qwen2024qwen25}, and Qwen3.5-4B, 9B, and 27B
\citep{qwen3.5}.  The benchmarks are
AIME 2024--2026 \citep{dekoninck2026matharena}, HMMT \citep{dekoninck2026matharena},
and MATH-100 \citep{hendrycks2021math}; the biology,
chemistry, and physics partitions of GPQA Diamond \citep{rein2023gpqa}, MMLU
Science \citep{hendrycks2020mmlu}, and SuperGPQA Science
\citep{du2025supergpqa}; a 100-question text-only HLE subset
\citep{phan2025hle}; and the easy, medium, and hard partitions of LiveCodeBench
\citep{jain2024livecodebench} restricted to problems released after March 30,
2024.  Together the groups contain 1,418 problems; their sizes are
\(N\in\{19,30,86,93,100\}\).

\textbf{Rollout corpus and task-probability profiles.}
For every problem--checkpoint combination, we independently sample \(K=1024\)
responses, yielding 8,712,192 responses.  Let \(Y_{ik}\in\{0,1\}\) indicate
whether rollout \(k\) on task \(i\) is correct under the benchmark-specific
scoring rule.  Appendix~\ref{app:rollout-generation-scoring} documents the
rollout and scoring protocol.  For each benchmark--checkpoint pair, we estimate
\(\widehat p_i^{(K)}=K^{-1}\sum_{k=1}^K Y_{ik}\).
Each of the 108 resulting vectors \(\widehat{\mathbf p}^{(K)}\) estimates the
task-probability profile of one benchmark--checkpoint pair.  The estimated
profile for GPQA Diamond Biology with Qwen2.5-3B is degenerate:
every \(\widehat p_i^{(K)}\) is zero or one, so evaluation under this fixed
empirical profile has zero variance and the Uniform-normalized ratio is
undefined.  We exclude this pair and retain
107 estimated task-probability profiles.

\subsection{Variance reduction and robustness}
\label{sec:variance-reduction}

We evaluate the statistical benefit of HBN through two questions:
\begingroup
\setlength{\leftmargini}{0.5\leftmargini}
\begin{itemize}
  \item \textit{RQ1: Does HBN reduce evaluation variance relative to Uniform
  across budgets?}
  \item \textit{RQ2: Do Bayesian regularization and hierarchical information
  sharing improve allocation?}
\end{itemize}
\endgroup

\paragraph{Experimental design.}
We fix the 107 retained task-probability profiles and evaluate budgets
\(b\in\{8,16,32,64\}\).  For each profile and candidate, we average 8192
independent pilot draws and compute continuation variance analytically using
\eqref{eq:two-stage-conditional-variance}.
The metric is Uniform-normalized variance, averaged equally across profiles
at each budget; Uniform corresponds to 100\%.  Oracle knows the profile and solves
the exact positive-integer Neyman problem, providing an infeasible
full-information benchmark rather than a deployable competitor.
Appendix~\ref{app:independent-analytic-replay} gives the replay and
random-seed details.

\paragraph{A1: Average variance gains increase with budget.}
Table~\ref{tab:bayesian-neyman-main} shows that HBN improves on Uniform at
every tested budget, with its mean normalized variance decreasing from
87.2\% at \(b=8\) to 66.4\% at \(b=64\).  Larger budgets therefore yield
greater average statistical benefit.  Oracle remains substantially lower,
at 40.7\%--39.3\%, indicating that full knowledge of task probabilities would
permit further gains; HBN must instead infer allocation scores from a
limited pilot.

HBN reduces variance relative to Uniform on nearly all profiles, but not
every profile.  Appendix~\ref{app:profile-robustness} reports the distribution
of HBN's variance gains and the single adverse profile, where HBN's variance
exceeds Uniform's by at most \(2.4\%\).

\paragraph{A2: Bayesian regularization and hierarchical sharing improve the observed allocations.}
To examine Bayesian regularization and hierarchical information sharing,
we compare HBN with two ablation variants, EN and IBN.  All three share the
same two-stage estimator and exact integer continuation allocator.  Each
variant receives its own pilot-and-weight design, since changing the
allocation scores also changes the balance between learning from the pilot
and allocating continuation samples.
\begingroup
\setlength{\leftmargini}{0.5\leftmargini}
\begin{itemize}
  \item \textbf{Tuned Empirical Neyman (EN)} removes Bayesian regularization
  and hierarchical information sharing, using the empirical pilot score
  \(S_i(m-S_i)/[m(m+1)]\).  It selects \((m,w)\) post hoc using the profiles
  with the same task count at each budget.
  \item \textbf{Tuned Independent Bayesian Neyman (IBN)} retains Bayesian
  regularization but removes hierarchical information sharing, using
  independent \(\operatorname{Beta}(\alpha,\alpha)\) priors.  It designs
  \((m,w)\) under each matched prior and selects only \(\alpha\) post hoc
  across profiles at each budget.
\end{itemize}
\endgroup
Both controls select their best-performing designs retrospectively
on the evaluation profiles, yielding hindsight envelopes over the searched
candidates.  This deliberately avoids tying the comparison to an arbitrary
default and gives the ablation variants an information advantage over HBN.
HBN fixes its reference hyperpriors
\(\xi,\delta\sim\operatorname{Beta}(1,1)\) and selects \((m,w)\) ex ante using
only \((N,b)\), without target evaluation outcomes.  Only its posterior and
continuation allocation adapt to the pilot.
Appendices~\ref{app:hbn-design-calculation} and
\ref{app:scores-exact-allocation} provide the HBN design and baseline
implementation details.

\begingroup
\setlength{\intextsep}{6pt plus 1pt minus 1pt}
\begin{table}[!htbp]
  \centering
  \small
  \setlength{\abovecaptionskip}{3pt}
  \setlength{\belowcaptionskip}{0pt}
  \setlength{\tabcolsep}{7pt}
  \begin{tabular}{@{}*{5}{>{\centering\arraybackslash}p{\dimexpr(0.8\textwidth-8\tabcolsep)/5\relax}}@{}}
    \toprule
    & Oracle & Tuned EN & Tuned IBN & HBN \\
    \midrule
    \(b=8\)  & 40.7\% & 94.3\% & 89.6\% & \textbf{87.2\%} \\
    \(b=16\) & 39.9\% & 92.6\% & 81.9\% & \textbf{80.1\%} \\
    \(b=32\) & 39.5\% & 91.9\% & 74.2\% & \textbf{73.0\%} \\
    \(b=64\) & 39.3\% & 91.5\% & 67.0\% & \textbf{66.4\%} \\
    \bottomrule
  \end{tabular}
  \caption{Mean Uniform-normalized variance across 107
  benchmark--checkpoint profiles (Uniform = 100\%); lower is better.
  EN and IBN use hindsight tuning; HBN uses ex-ante design.}
  \label{tab:bayesian-neyman-main}
\end{table}
\endgroup

Table~\ref{tab:bayesian-neyman-main} shows successively lower mean variance
from Tuned EN to Tuned IBN to HBN at every budget.  HBN outperforms both
hindsight envelopes despite committing to its design ex ante.

Even with hindsight tuning, EN reduces variance by only
\(5.7\%\)--\(8.5\%\).  Its score is zero for all-failure or all-success
pilots, although the true task variance may be positive.  When other tasks
have positive scores, such tasks typically receive only the mandatory one
fresh rollout.  Longer pilots reduce false zeros but shrink the continuation
budget; tuning \((m,w)\) cannot eliminate this tradeoff.

IBN's positive \(\alpha\) prevents zero scores, alleviating this failure
and yielding normalized variances of \(89.6\%\)--\(67.0\%\).  However, its
independent priors do not share pilot information across tasks: tuning one
\(\alpha\) per budget adjusts overall shrinkage, not each profile's latent
mean and dispersion.  HBN infers these shared quantities from the complete
pilot, allowing tasks to borrow strength from one another.  This mechanism
is consistent with its advantage, although the comparison evaluates complete
designs rather than isolating information sharing alone.

Beyond these ablations, we adapt TS-Neyman \citep{morikawa2026tsneyman}
as an external baseline for our Bernoulli setting.  HBN achieves lower mean
Uniform-normalized MSE at all four budgets, even with hindsight tuning of
this baseline (Appendix~\ref{app:ts-neyman}).

\subsection{Computation, system overhead, and practical benefit}
\label{sec:system-overhead}

The variance gains in Section~\ref{sec:variance-reduction} must be weighed
against the cost of generating a non-uniform workload and waiting for pilot
feedback.  The systems question is whether a concrete asynchronous implementation
can reduce the pilot barrier sufficiently for the statistical gains to remain
beneficial after accounting for execution costs.  Using the HBN-sync and HBN-async schemes defined in
Section~\ref{sec:partial-allocation-prefetch}, we study three questions:
\begingroup
\setlength{\leftmargini}{0.5\leftmargini}
\begin{itemize}
  \item \textit{RQ3: What additional computation does HBN introduce?}
  \item \textit{RQ4: Does HBN-async mitigate the impact of the pilot
  synchronization point through speculative pre-allocation?}
  \item \textit{RQ5: When does speculative evaluation offer a clear
  practical benefit?}
\end{itemize}
\endgroup

\paragraph{Experimental design.}
We use the same 107 task-probability profiles and budgets
\(b\in\{8,16,32,64\}\), giving 428 profile--budget settings.  Generation
runs on eight H20 GPUs per worker, with a shared concurrency limit of
\(C=256\) outstanding requests (32 per replica) for all three strategies.
We use the frozen prompts and model settings of the rollout corpus.
To compare synchronous and asynchronous execution under the same
statistical allocation, we couple their pilot rewards.  Using generated-answer
rewards would generally produce different pilot outcomes and hence different
continuation allocations across runs, mixing execution-policy effects with
allocation changes.  We therefore assign each logical request ID a fixed
Bernoulli reward with probability \(\widehat p_i^{(K)}\), shared across
execution strategies and revealed only when its real generation completes.
HBN-sync and HBN-async consequently receive identical pilot rewards and
select the same final continuation IDs.  This controls the allocation
decision while retaining real generation and execution costs; realized
response lengths may still differ across runs.  Benchmark judging and
code-execution latency are excluded.

\emph{Uniform} makes all \(Nb\) uniformly allocated requests eligible from
the start.  We compare it with HBN-sync and HBN-async under this common
concurrency limit.
Each strategy has one successful measured execution
per profile--budget setting, for 1,284 policy executions.
We summarize execution costs across the 107 profiles at each budget,
targeting aggregate behavior over the evaluated workload suite.  These
measurements do not quantify repeated-run variability within an individual
setting.
We additionally measure 428 Uniform executions at rounded-down
variance-equivalent budgets for the benefit--cost comparison below.
Appendix~\ref{app:systems-implementation} gives the execution and accounting
details.

\paragraph{A3: Useful computation and wasted speculative computation.}
HBN's computation relative to Uniform can be decomposed into two parts.
We distinguish estimated \emph{useful FLOPs}, \(F^{\mathrm{useful}}\), for
accepted rollouts from \emph{wasted FLOPs}, \(F^{\mathrm{waste}}\), for
discarded or running-aborted requests.  Their sum is the estimated actual
computation, denoted by \(F\).
This decomposition applies to both HBN-sync and HBN-async.
The first is the change in useful computation associated with task allocation:
HBN changes the distribution
of accepted rollouts across tasks, whose generation costs differ, while
preserving the total count \(Nb\).  The second, present in HBN-async, is wasted
speculative computation from pre-allocated rollouts that are not selected by the final
allocation, including completed-discarded and running-aborted requests.
Since Uniform's actual computation is entirely useful, the accounting identity is
\begin{equation}
 F_{\mathrm{HBN}}-F_{\mathrm{Uniform}}
 =\bigl(F^{\mathrm{useful}}_{\mathrm{HBN}}-F_{\mathrm{Uniform}}\bigr)
  +F^{\mathrm{waste}}_{\mathrm{HBN}}.
 \label{eq:systems-compute-decomposition}
\end{equation}
Here HBN denotes either execution scheme; for HBN-sync,
\(F^{\mathrm{waste}}_{\mathrm{HBN}}=0\).
Table~\ref{tab:systems-compute} reports this decomposition for HBN-async.
Useful computation increases by 2.00\%--5.62\% in aggregate over Uniform,
whereas wasted speculative computation falls from 2.23\% of Uniform's
computation at \(b=8\) to 0.026\% at \(b=64\).  Thus, at larger budgets,
the additional computation is predominantly in accepted generations, even though
the number of accepted rollouts is identical.

\begingroup
\setlength{\intextsep}{6pt plus 1pt minus 1pt}
\begin{table}[!htbp]
  \centering
  \small
  \setlength{\abovecaptionskip}{3pt}
  \setlength{\belowcaptionskip}{0pt}
  \setlength{\tabcolsep}{4pt}
  \begin{tabular}{@{}rrrrr@{}}
    \toprule
    \(b\) & Extra rollouts & Useful FLOPs & Wasted FLOPs & Actual FLOPs \\
          & vs. Uniform & vs. Uniform & vs. Uniform & vs. Uniform \\
    \midrule
     8 & +2.89\%  & +2.00\% & +2.23\%  & +4.23\% \\
    16 & +0.98\%  & +3.03\% & +0.51\%  & +3.54\% \\
    32 & +0.25\%  & +4.41\% & +0.073\% & +4.49\% \\
    64 & +0.077\% & +5.62\% & +0.026\% & +5.64\% \\
    \bottomrule
  \end{tabular}
  \caption{HBN-async computation across 107 profiles at each budget.
  Extra rollouts report the relative increase in actual rollout count over Uniform.
  FLOPs are estimated using the method in Appendix~\ref{app:systems-implementation}.
  The Useful, Wasted, and Actual FLOP columns correspond to the first
  right-hand term, second right-hand term, and left-hand side of
  Eq.~\eqref{eq:systems-compute-decomposition}, respectively, each summed
  across profiles at each budget and divided by summed Uniform FLOPs.}
  \label{tab:systems-compute}
\end{table}
\endgroup

\paragraph{A4: Pre-allocation reduces the observed execution overhead.}
Despite its wasted FLOPs, HBN-async reduces both actual time and
useful-FLOP-normalized time relative to HBN-sync at every tested budget.

\begingroup
\setlength{\intextsep}{6pt plus 1pt minus 1pt}
\begin{table}[!htbp]
  \centering
  \small
  \setlength{\abovecaptionskip}{3pt}
  \setlength{\belowcaptionskip}{0pt}
  \setlength{\tabcolsep}{4pt}
  \begin{tabular}{@{}>{\centering\arraybackslash}p{1.5em}
    *{6}{>{\centering\arraybackslash}p{\dimexpr(\linewidth-1.5em-12\tabcolsep)/6\relax}}@{}}
    \toprule
    & \multicolumn{3}{c}{Actual time vs. Uniform}
    & \multicolumn{3}{c}{Useful-FLOP-normalized time vs. Uniform} \\
    \cmidrule(lr){2-4}\cmidrule(lr){5-7}
    \(b\) & HBN-sync & HBN-async & \(\Delta\)
          & HBN-sync & HBN-async & \(\Delta\) \\
    \midrule
     8 & +33.73\% & +11.15\% & $22.57\%$ & +30.48\% & +8.45\% & $22.03\%$ \\
    16 & +16.47\% &  +2.85\% & $13.61\%$ & +13.39\% & +0.14\% & $13.25\%$ \\
    32 & +13.38\% &  +5.49\% &  $7.89\%$ &  +9.27\% & +1.33\% &  $7.94\%$ \\
    64 & +11.10\% &  +6.56\% &  $4.54\%$ &  +5.61\% & +1.38\% &  $4.23\%$ \\
    \bottomrule
  \end{tabular}

  \caption{System time across 107 profiles at each budget.  Strategy columns report
  percentage increases over Uniform, computed from ratios of total times
  across all 107 profiles at the same budget, with useful-FLOP normalization
  applied per profile where indicated.  \(\Delta\) denotes the
  HBN-sync--HBN-async difference in percentage points.}
  \label{tab:systems-time}
\end{table}
\endgroup

To better assess the execution overhead associated with the pilot barrier,
we approximately adjust runtime for differences in useful computation.
Otherwise, a time difference can reflect a different amount
of accepted generation rather than a difference in execution efficiency.
For strategy \(p\) and profile \(j\) at a fixed budget, \emph{useful-FLOP-normalized time}
is \(\widetilde T_{p,j}=T_{p,j}F^{\mathrm{useful}}_{\mathrm{Uniform},j}/F^{\mathrm{useful}}_{p,j}\):
the measured time rescaled to Uniform's useful computation in that setting.
We normalize each profile--budget setting, then average equally over the
107 profiles at each budget.  Wasted computation remains included in runtime
but does not count toward useful FLOPs in this adjustment.

Table~\ref{tab:systems-time} shows that HBN-async reduces both actual and
useful-FLOP-normalized overhead relative to HBN-sync at every budget, despite
speculative waste.  Similar \(\Delta\) values before and after normalization
indicate that the reduction persists after accounting for useful computation.
The gap is largest at \(b=8\) and narrows with budget.

For \(b\geq16\), HBN-async's mean useful-FLOP-normalized time is less than
1.4\% above Uniform's.
Thus, under this hardware and concurrency configuration, speculative
execution reduces the remaining normalized system overhead to a marginal
level at the larger budgets.  At \(b=8\), the residual overhead is still
8.45\%.  Intuitively, at small budgets, waiting for the final pilot rewards
can account for a substantial fraction of runtime, with relatively little
continuation computation available to hide that wait.  Larger budgets
provide more continuation computation to overlap with the pilot and spread
the remaining synchronization cost over a longer run, leaving a small
relative overhead.

\paragraph{A5: Sufficient statistical gain and a total rollout budget well above the concurrency limit.}
Let \(r_{j,b}\) be HBN's Uniform-normalized variance from the replay in
Table~\ref{tab:bayesian-neyman-main}.  By Uniform's \(1/b\) variance scaling,
its \emph{continuous-equivalent budget}, \(b/r_{j,b}\), matches HBN's variance
under these replay ratios.  Its \emph{rounded-equivalent budget},
\(b^{\downarrow}_{j,b}=\lfloor b/r_{j,b}\rfloor\), rounds this value down
to an integer for execution, leaving Uniform with variance no lower than HBN's.
We freshly measure Uniform at this budget with unchanged generation and
concurrency settings.

\begingroup
\setlength{\intextsep}{6pt plus 1pt minus 1pt}
\begin{table}[!htbp]
  \centering
  \small
  \setlength{\abovecaptionskip}{3pt}
  \setlength{\belowcaptionskip}{0pt}
  \setlength{\tabcolsep}{4pt}
  \begin{tabular*}{\textwidth}{@{\extracolsep{\fill}}ccccccc@{}}
    \toprule
    & \multicolumn{3}{c}{Equivalent rollout budget} & \multicolumn{3}{c}{Actual time vs. Uniform} \\
    \cmidrule(lr){2-4}\cmidrule(lr){5-7}
    \(b\) & \begin{tabular}[c]{@{}c@{}}Avg.\\continuous\end{tabular}
          & \begin{tabular}[c]{@{}c@{}}Avg.\\rounded\end{tabular}
          & \begin{tabular}[c]{@{}c@{}}Rounded /\\continuous\end{tabular}
          & \begin{tabular}[c]{@{}c@{}}Rounded-equivalent\\Uniform\end{tabular}
          & HBN-async
          & \(\Delta\) \\
    \midrule
     8 & 9.37 & 8.87 & 94.63\% &  +3.79\% & +11.15\% & $-7.36\%$ \\
    16 & 20.88 & 20.39 & 97.66\% & +13.64\% &  +2.85\% & $+10.79\%$ \\
    32 & 47.04 & 46.56 & 98.98\% & +27.95\% &  +5.49\% & $+22.46\%$ \\
    64 & 107.22 & 106.75 & 99.56\% & +42.81\% &  +6.56\% & $+36.25\%$ \\
    \bottomrule
  \end{tabular*}
  \caption{Measured benefit--cost comparison across 107 profiles.
  Average budgets are task-count-weighted.
  The budget ratio reports the fraction of continuous-equivalent rollouts
  actually executed after rounding down:
  \(\sum_j N_j b^{\downarrow}_{j,b}/\sum_j N_j(b/r_{j,b})\), as a percentage.
  Time increases use summed original-budget Uniform time as the denominator;
  HBN-async values are from Table~\ref{tab:systems-time}.
  \(\Delta\) is rounded-equivalent Uniform minus HBN-async, in percentage points.}
  \label{tab:systems-benefit}
\end{table}
\endgroup

At \(b\geq16\), rounded-equivalent Uniform takes 13.64\%--42.81\% more
total time than original-budget Uniform, versus 2.85\%--6.56\% for
HBN-async (Table~\ref{tab:systems-benefit}).  HBN-async thus retains a
clear aggregate time advantage despite rounding in Uniform's favor.
Rounding removes 5.37\% of
the continuous rollout budget at \(b=8\), falling to 0.44\% at \(b=64\).

At \(b=8\), HBN-async is slower than rounded-equivalent Uniform.
Two factors limit its benefit.  First, the statistical gain is smaller,
and rounding further reduces the comparator's additional budget: it retains
only 63.36\% of the extra rollouts prescribed by continuous variance matching.
Second, the smaller total budget \(Nb\) provides less continuation computation
relative to the concurrency limit \(C=256\), leaving less opportunity to
overlap generation with the pilot barrier.  HBN-async consequently retains
8.45\% useful-FLOP-normalized overhead (Table~\ref{tab:systems-time}).
These effects leave no clear overall benefit in this comparison.

The widening margin in Table~\ref{tab:systems-benefit} reflects a combination
of statistical and systems effects.  Statistically, the opportunity for
nonuniform allocation depends on the task-probability profile, while larger
budgets allow HBN to exploit that opportunity more effectively, yielding
greater average variance reductions in our experiments.  On the systems
side, increasing \(b\) raises the total rollout budget \(Nb\) relative to
the fixed concurrency limit \(C\), providing more continuation computation
to overlap with the pilot and reducing the relative impact of residual
barrier delays.  Speculative evaluation is therefore most beneficial when
\textbf{profiles offer substantial allocative opportunity} and
\textbf{the budget is sufficient both to exploit it statistically and to
amortize synchronization overhead}.
Under our experimental setup, this comparison indicates a clear benefit
at \(b\geq16\).

\section{Conclusion and Limitations}
\label{sec:conclusion}

Speculative Evaluation combines hierarchical Bayesian allocation and
asynchronous pre-allocation to improve stochastic LLM evaluation.  HBN
reduces variance through hierarchical information sharing and ex-ante
pilot-and-weight design; HBN-async mitigates the pilot barrier despite
speculative waste.  Across 107 benchmark--checkpoint profiles, their
combination yields clear aggregate benefits at \(b\geq16\) under our setup.
Several limitations remain:
\begingroup
\setlength{\leftmargini}{0.5\leftmargini}
\begin{itemize}
  \setlength{\itemsep}{1pt}
  \setlength{\parsep}{0pt}
  \item \textbf{Statistical scope.} We assume conditionally independent
  Bernoulli outcomes.  Profile structure and prior misspecification affect
  efficiency, and gains are not guaranteed for every profile.
  \item \textbf{Systems scope.} Coupled synthetic rewards control the final
  allocation across HBN execution schemes.  They do not reproduce within-task
  dependence between correctness and generation latency, and the timing
  excludes benchmark scoring.  Practical benefits depend on generation and
  scoring costs as well as concurrency.
  \item \textbf{Timing precision.} Given the cost of real generation, each
  policy is measured once per profile--budget setting.  The reported time
  benefits are aggregate observations under the tested hardware and
  concurrency; repeated-run variability within individual settings remains
  unquantified.
\end{itemize}
\endgroup
Future work could extend the framework to nonbinary or correlated outcomes
and develop cost-aware allocation with heterogeneous generation costs and
real scoring feedback.

\clearpage
\bibliographystyle{iclr2027_conference}
\bibliography{references}

\clearpage
\appendix
\section{Complete Algorithm}
\label{app:complete-algorithm}

\begin{algorithm}[H]
  \caption{Speculative evaluation with Hierarchical Bayesian Neyman (HBN)
  allocation}
  \label{alg:bayesian-neyman}
  \footnotesize
  \begin{algorithmic}[1]
    \Require Tasks \(1{:}N\), integer budget \(b\geq2\) per task, reference hyperprior
      \(\pi_\xi=\pi_\delta=\operatorname{Beta}(1,1)\), and rollout procedure
      \(\Call{Rollout}{i}\)
    \Ensure Unbiased estimate \(\widehat\mu_w\)
    \Statex \Comment{\textsc{Offline design}}
    \State Jointly choose \((m,w)\gets(m_{\mathrm B}^*,w_{\mathrm B}^*)\) by
      \eqref{eq:bayes-optimal-stage-weight} and minimizing over feasible pilot sizes
      \Comment{uses no target outcomes}
    \Statex \Comment{\textsc{Stage 1: uniform pilot}}
    \State In parallel, draw \(Y^{(0)}_{i,1:m}\gets\Call{Rollout}{i}\) and set
      \(S_i\gets\sum_{t=1}^mY^{(0)}_{i,t}\) for all \(i\)
    \Statex \Comment{\textsc{Posterior update and exact integer allocation}}
    \State \textbf{synchronize}; compute \(\pi(\xi,\delta\mid S)\) by
      \eqref{eq:hyperposterior} and \(\widetilde v_i(S)\) by
      \eqref{eq:posterior-rollout-variance}
      \Comment{only synchronization}
    \State Initialize \(L_i\gets1\) for all \(i\)
      \Comment{guarantees a fresh rollout per task}
    \While{\(\sum_iL_i<N(b-m)\)}
      \State \(k\gets\arg\max_i
        \widetilde v_i(S)/[L_i(L_i+1)]\); \(L_k\gets L_k+1\)
        \Comment{largest marginal reduction}
    \EndWhile
    \Statex \Comment{\textsc{Stage 2: adaptive continuation}}
    \State In parallel, draw fresh
      \(Y^{(1)}_{i,1:L_i}\gets\Call{Rollout}{i}\) for all \(i\)
    \Statex \Comment{\textsc{Unbiased aggregation}}
    \State Form \(\widehat\mu_0,\widehat\mu_1\) from both stages and
      \Return \(\widehat\mu_w\) in
      \eqref{eq:two-stage-estimator}
      \Comment{recycles the pilot}
  \end{algorithmic}
\end{algorithm}

\begin{algorithm}[H]
  \caption{HBN-async: speculative pre-allocation and fixed-ID recovery}
  \label{alg:hbn-async}
  \footnotesize
  \begin{algorithmic}[1]
    \Require Tasks \(1{:}N\), integer budget \(b\geq2\) per task, reference prior \(\pi\),
      concurrency limit \(C\), and rollout procedure \(\Call{Rollout}{i}\)
    \Ensure Estimate \(\widehat\mu_w\) using the final HBN allocation
    \Statex \Comment{\textsc{Offline design and pilot submission}}
    \State Fix \((m,w)\) as in Algorithm~\ref{alg:bayesian-neyman}
    \State Fix pilot IDs and continuation IDs \((i,\ell)\) before execution
    \State Initialize scored pilot counts \(n_i\gets0\), successes \(S_i\gets0\), and result store
    \State Submit all pilot requests before any continuation request
      \Comment{at most \(C\) active requests}
    \Statex \Comment{\textsc{Partial-pilot speculation}}
    \While{some pilot reward is unavailable}
      \State Receive a completion and store its result by its designated ID
      \If{the result is a pilot reward for task \(i\)}
        \State Update \(n_i\) and \(S_i\)
        \If{all pilot rewards are available}
          \State \textsc{break}
        \EndIf
        \State Compute partial scores \(\widetilde v_i^{(t)}\) by
          \eqref{eq:partial-posterior-rollout-variance}
        \State Compute \((L_i^{\mathrm B,(t)})_{i=1}^N\) with the exact integer allocator
        \State Refresh pending continuations to unsubmitted IDs with
          \(\ell\leq L_i^{\mathrm B,(t)}\)
        \State Order them by increasing \(\ell/L_i^{\mathrm B,(t)}\), with deterministic ties
        \State Leave running and completed continuations unchanged
      \EndIf
      \State Fill available capacity with pending pilots first, then eligible continuations
    \EndWhile
    \Statex \Comment{\textsc{Final allocation and recovery}}
    \State Compute \((L_i^{\mathrm B}(S))_{i=1}^N\) from the complete pilot by
      \eqref{eq:integer-bayesian-neyman}
    \State Set required continuation IDs
      \(\mathcal I\gets\{(i,\ell):1\leq\ell\leq L_i^{\mathrm B}(S)\}\)
    \State Retain completed results and running requests in \(\mathcal I\)
    \State Discard completed surplus, cancel queued surplus, and abort running surplus
    \State Submit missing IDs in \(\mathcal I\), respecting the concurrency limit
    \State Await every required ID; never substitute an earlier-finishing surplus result
    \Statex \Comment{\textsc{Aggregation}}
    \State Form \(\widehat\mu_0\) from pilot results and \(\widehat\mu_1\) from IDs in \(\mathcal I\)
    \State \Return \(\widehat\mu_w=w\widehat\mu_0+(1-w)\widehat\mu_1\)
  \end{algorithmic}
\end{algorithm}

\section{Experimental and Implementation Details}
\label{app:experimental-details}

\subsection{Rollout generation and scoring}
\label{app:rollout-generation-scoring}

We generate every response with vLLM 0.23.0 using temperature \(0.6\),
top-\(p=0.95\), no top-\(k\) truncation, a 16,384-token cap, and bfloat16
inference.  The model context
limit is 32,768 tokens and prefix caching is disabled.  Qwen2.5 runs in
non-thinking mode; Qwen3.5 uses thinking mode with at most 8,192 thinking
tokens within the total cap.  We use master seed 42 with deterministic
per-request seeds.

Generation workers each contain eight NVIDIA H20 GPUs with 96 GB per GPU.
Each GPU hosts one persistent tensor-parallel-size-one replica, while a routing
process dynamically distributes requests among the eight replicas.
The GPU-memory-utilization limit is 0.9.  The engine sequence limits for
Qwen2.5-3B/7B/32B and Qwen3.5-4B/9B/27B are respectively
256/192/128 and 128/128/64.

Prompts, extraction, and scoring are benchmark-specific: mathematical tasks
use normalized numeric or symbolic matching, and multiple-choice tasks use
option extraction.  HLE uses the CAIS judge prompt with a fixed
Qwen3.6-35B-A3B checkpoint in non-thinking mode at temperature zero.
LiveCodeBench programs are evaluated against test cases on separate CPU
workers in network-isolated, resource-limited environments.

Generation of the complete 108-pair corpus required approximately 411 active
model-job hours, equivalent to 17.1 days if all model-level jobs were run
sequentially on one independent eight-H20 worker.  This estimate excludes
scoring and is not an exact GPU-hour count because some jobs used multiple
eight-GPU workers concurrently.

\subsection{HBN design and numerical implementation}
\label{app:hbn-design-calculation}

For each candidate pilot, HBN estimates the continuation mass \(B_{\pi,m}\) in
\eqref{eq:prior-variance-masses}, obtains the optimal weight in closed form from
\eqref{eq:bayes-optimal-stage-weight}, and compares the resulting Bayes risks.
It evaluates the full feasible grid \(m=1,\ldots,b-1\); every selected solution
is interior to that search range.  It averages eight independent 8192-draw replicates under
its hierarchical reference prior, using common random numbers across candidate
pilots within each replicate.  Its tensor-product Gauss--Legendre quadrature
has order 16 in each of \(\xi\) and \(\delta\).
The same quadrature grid is used for complete- and partial-pilot posterior
updates online.  Posterior weights are computed in the log domain using
log-beta evaluations and log-sum-exp normalization.
Table~\ref{tab:hbn-ex-ante-schedule} reports the resulting ex-ante schedules.

\begin{table}[H]
  \centering
  \small
  \setlength{\abovecaptionskip}{3pt}
  \setlength{\belowcaptionskip}{0pt}
  \setlength{\tabcolsep}{5.5pt}
  \begin{tabular}{rcccc}
    \toprule
    \(N\) & \(b=8\) & \(b=16\) & \(b=32\) & \(b=64\) \\
    \midrule
    19  & \((4,0.49)\) & \((6,0.35)\) & \((10,0.29)\) & \((17,0.24)\) \\
    30  & \((4,0.48)\) & \((6,0.35)\) & \((10,0.29)\) & \((16,0.22)\) \\
    86  & \((4,0.48)\) & \((6,0.35)\) & \((10,0.28)\) & \((16,0.22)\) \\
    93  & \((4,0.48)\) & \((6,0.35)\) & \((10,0.29)\) & \((16,0.22)\) \\
    100 & \((4,0.48)\) & \((6,0.35)\) & \((10,0.28)\) & \((16,0.22)\) \\
    \bottomrule
  \end{tabular}
  \caption{Ex-ante HBN design selected under the reference prior.  Each cell
  reports \((m_{\mathrm B}^*,w_{\mathrm B}^*)\) for task count \(N\) and
  average budget \(b\).  We display weights to two decimal places; all
  experiments use their full-precision values.  No target-pair outcome enters
  the selection.}
  \label{tab:hbn-ex-ante-schedule}
\end{table}

\subsection{Baseline scores and tuning}
\label{app:scores-exact-allocation}
\label{app:en-ibn-tuning}

For \(\alpha>0\), Independent Bayesian Neyman (IBN) uses
\begin{equation}
  \widetilde v_i^{\mathrm{IBN}}
  =\frac{(S_i+\alpha)(m-S_i+\alpha)}
         {(m+2\alpha)(m+2\alpha+1)}.
  \label{eq:ibn-score}
\end{equation}
This is the posterior expectation of \(p_i(1-p_i)\) under the matched prior
\(p_i\stackrel{\mathrm{iid}}{\sim}\operatorname{Beta}(\alpha,\alpha)\).
The \(\alpha\downarrow0\) allocation-rule limit is Empirical Neyman (EN):
\begin{equation}
  \widetilde v_i^{\mathrm{EN}}
  =\frac{S_i(m-S_i)}{m(m+1)}.
  \label{eq:en-score}
\end{equation}
The factor \(m/(m+1)\) relative to the plug-in Bernoulli variance is common to
all tasks and therefore does not change the allocation.

Every method passes its nonnegative scores to the decreasing-marginal-gain
allocator in Section~\ref{sec:posterior-neyman-allocation}.  It starts at one continuation
rollout per task and assigns the remaining budget exactly.  If all EN scores
are zero, its continuation is uniform.

EN uses the evaluation profiles to select \((m,w)\) for each
\((N,b)\).  IBN designs \((m,w)\) under each candidate prior, then uses the
profiles to select \(\alpha\) at each budget.

\paragraph{EN design pooled over task-probability profiles.}
Index the retained task-probability profiles by \(j\), and let
\(\mathcal J_N=\{j:N_j=N\}\).  The five task counts
\(N\in\{19,30,86,93,100\}\) contain respectively
\(5,24,6,6,66\) task-probability profiles.  For task-probability profile \(j\), write
\(v_{ji}=\widehat p_{ji}^{(K)}(1-\widehat p_{ji}^{(K)})\) and
\(V_j=\sum_i v_{ji}\).  For every \(b\) and every
\(m=1,\ldots,b-1\), we draw 8192 pilot vectors from the fixed
task-probability profile, form the EN scores in \eqref{eq:en-score}, and compute the exact
continuation allocation \(\mathbf L_j^{\mathrm{EN}}(S;m,b)\).  This estimates
\begin{equation}
  C_{j,b,m}^{\mathrm{EN}}
  =\mathbb E_{S\mid\widehat{\mathbf p}_j^{(K)}}\!\left[
    \frac{1}{V_j}\sum_{i=1}^{N}
    \frac{v_{ji}}{L_{ji}^{\mathrm{EN}}(S;m,b)}
  \right],
  \qquad
  \overline C_{N,b,m}^{\mathrm{EN}}
  =\frac{1}{|\mathcal J_N|}\sum_{j\in\mathcal J_N}
    C_{j,b,m}^{\mathrm{EN}}.
  \label{eq:en-tuning-coefficient}
\end{equation}
Because the pilot contribution to each task-probability profile's Uniform-normalized variance
is \(b w^2/m\), the group-mean candidate objective is
\begin{equation}
  \mathcal R_{N,b}^{\mathrm{EN}}(m,w)
  =b\left\{\frac{w^2}{m}
    +(1-w)^2\overline C_{N,b,m}^{\mathrm{EN}}\right\}.
  \label{eq:en-tuning-risk}
\end{equation}
No weight-grid search is needed.  For each candidate \(m\), the minimizing
weight and risk are
\begin{equation}
  w_{N,b,m}^{\mathrm{EN}}
  =\frac{\overline C_{N,b,m}^{\mathrm{EN}}}
  {m^{-1}+\overline C_{N,b,m}^{\mathrm{EN}}},
  \qquad
  \mathcal R_{N,b}^{\mathrm{EN},*}(m)
  =b\frac{m^{-1}\overline C_{N,b,m}^{\mathrm{EN}}}
  {m^{-1}+\overline C_{N,b,m}^{\mathrm{EN}}},
  \label{eq:en-tuning-closed-form}
\end{equation}
The minimizing integer \(m\) is then selected, breaking an exact tie toward
the smaller pilot.  This produces one \((m,w)\) schedule for each \((N,b)\).

Table~\ref{tab:en-tuned-schedule} reports the selected schedules.
\begin{table}[H]
  \centering
  \small
  \begin{tabular}{rcccc}
    \toprule
    \(N\) & \(b=8\) & \(b=16\) & \(b=32\) & \(b=64\) \\
    \midrule
    19  & \((6,0.718)\) & \((14,0.814)\) & \((27,0.764)\) & \((53,0.728)\) \\
    30  & \((6,0.726)\) & \((14,0.835)\) & \((29,0.869)\) & \((59,0.887)\) \\
    86  & \((6,0.685)\) & \((13,0.722)\) & \((27,0.740)\) & \((54,0.735)\) \\
    93  & \((6,0.710)\) & \((13,0.755)\) & \((27,0.771)\) & \((51,0.707)\) \\
    100 & \((6,0.702)\) & \((14,0.804)\) & \((28,0.797)\) & \((55,0.779)\) \\
    \bottomrule
  \end{tabular}
  \caption{Tuned EN's post-hoc selected \((m,w)\) for each task count and
  budget.  Weights are displayed to three decimal places; replay uses
  full-precision values.}
  \label{tab:en-tuned-schedule}
\end{table}

\paragraph{IBN inner matched-prior design.}
The positive prior-strength grid is
\[
  \mathcal A=\{0.01,0.02,\ldots,0.20,0.25,0.50,1.00\}.
\]
For each \(\alpha\in\mathcal A\), the inner design treats
\(p_i\stackrel{\mathrm{iid}}{\sim}\operatorname{Beta}(\alpha,\alpha)\), for
which
\(\overline v_\alpha=\mathbb E[p_i(1-p_i)]
=\alpha/[2(2\alpha+1)]\).  For every \((\alpha,N,b,m)\), with
\(m=1,\ldots,b-1\), it estimates
\begin{equation}
  A_{\alpha,N,m}=\frac{N\overline v_\alpha}{m},
  \qquad
  B_{\alpha,N,b,m}
  =\mathbb E_{S\sim\pi_\alpha}\!\left[
    \sum_{i=1}^{N}
    \frac{\widetilde v_i^{\mathrm{IBN}}(S)}
         {L_i^{\mathrm{IBN}}(S;m,b)}
  \right].
  \label{eq:ibn-inner-masses}
\end{equation}
The expectation is evaluated with eight independent replicates of 8192
prior-predictive panels.  We generate the Beta--Bernoulli pilot paths through
the equivalent P\'olya-urn predictive law, thereby integrating over both
\(\mathbf p\) and \(S\).  Within a replicate, common uniforms give nested paths
across pilot sizes and are also reused across \(\alpha\) values.

As with HBN, the matched-prior weight and risk are then computed in closed form:
\begin{equation}
  w_{\alpha,N,b,m}^{\mathrm{IBN}}
  =\frac{B_{\alpha,N,b,m}}
         {A_{\alpha,N,m}+B_{\alpha,N,b,m}},
  \qquad
  \mathcal R_{\alpha,N,b}^{\mathrm{IBN},*}(m)
  =\frac{b}{N\overline v_\alpha}
   \frac{A_{\alpha,N,m}B_{\alpha,N,b,m}}
        {A_{\alpha,N,m}+B_{\alpha,N,b,m}}.
  \label{eq:ibn-inner-risk}
\end{equation}
For every \((\alpha,N,b)\), we select the risk-minimizing integer \(m\), again
breaking exact ties toward smaller \(m\), and retain its closed-form weight.

\paragraph{IBN outer hindsight selection.}
Each of the 23 matched-prior schedules is next replayed on every retained
task-probability profile.  For task-probability profile \(j\), budget \(b\), and strength \(\alpha\), let
\(\widehat r_{j,b,\alpha}\) be the Uniform-normalized variance averaged over
8192 independent pilot vectors.  The displayed IBN entry selects separately
at each budget
\begin{equation}
  \alpha_b^*
  \in\arg\min_{\alpha\in\mathcal A}
  \widehat r_{b,\alpha},
  \qquad
  \widehat r_{b,\alpha}
  =\frac{1}{107}\sum_{j=1}^{107}\widehat r_{j,b,\alpha},
  \label{eq:ibn-outer-tuning}
\end{equation}
with an exact tie broken toward smaller \(\alpha\).

The selected strengths are \(\alpha_b^*=0.13,0.13,0.12,0.11\) at
\(b=8,16,32,64\), respectively.  These are the IBN settings used in
Table~\ref{tab:bayesian-neyman-main}.

\subsection{Variance evaluation protocol}
\label{app:independent-analytic-replay}

Design and replay use distinct deterministic random-seed namespaces.  For each
benchmark--checkpoint pair, budget, and candidate policy, replay averages 8192
independent pilot vectors.  It then evaluates the continuation contribution
analytically via \eqref{eq:two-stage-conditional-variance}; no second-stage
Bernoulli outcomes are simulated.  EN candidate pilots share nested paths
during design.  During IBN replay, all \(\alpha\) candidates for a fixed
task-probability profile and budget share one nested Bernoulli path through the largest selected
pilot, reducing Monte Carlo noise in the outer comparison.  Allocation ties
use task-index order; an all-zero EN score vector uses uniform continuation.
All 107 retained task-probability profiles receive equal weight in
Table~\ref{tab:bayesian-neyman-main}.

\subsection{System execution and cost accounting}
\label{app:systems-implementation}

\paragraph{Frozen workload and reward protocol.}
The system study in Section~\ref{sec:system-overhead} uses the 107 retained
profiles, all four budgets, and the full-precision ex-ante HBN designs in
Table~\ref{tab:hbn-ex-ante-schedule}.  To compare execution schemes under
the same statistical allocation, we couple their pilot rewards: rewards from
generated answers could otherwise change the pilot and final allocation
across runs.  Bernoulli rewards with probability \(\widehat p_i^{(K)}\) are
fixed by logical request ID and shared across execution strategies.  They
are revealed only after real generation completes.  This gives HBN-sync and
HBN-async identical pilot rewards and final continuation IDs while measuring
real generation time, excluding benchmark scoring latency.
Useful-FLOP differences reflect both task allocation and schedule-dependent
response lengths.

\paragraph{Serving and timing.}
We use the generation configuration in Appendix~\ref{app:rollout-generation-scoring},
with eight persistent single-GPU replicas per worker.
The common router admits at most 32
outstanding requests per replica, for a total window of 256.  It assigns
eligible requests to the least-loaded replica and replenishes capacity as
completion events arrive.

The main timer starts before initial dispatch and ends when all formally
selected request IDs have completed and their rewards are available.
Pilot-dependent posterior inference, integer allocation, and online queue
management are included.  Model loading, warmup, ex-ante design, static
quadrature setup, and between-policy reset are excluded.  Between policies,
we drain outstanding work and clear request and cache state while retaining
the loaded models.

\paragraph{Execution policies.}
The execution policies follow
Section~\ref{sec:partial-allocation-prefetch} and Algorithm~\ref{alg:hbn-async}.
HBN-async computes provisional allocations in the background, starting from
the prior and updating from the latest available partial-pilot snapshot;
intermediate snapshots may be skipped while generation proceeds.
The exact integer allocator uses threshold search followed by marginal-gain
corrections.

\paragraph{Rounded-equivalent Uniform measurements.}
For each of the 428 profile--budget settings, we derive
\(b^{\downarrow}_{j,b}=\lfloor b/r_{j,b}\rfloor\) from the unrounded HBN
replay ratio and perform a fresh Uniform execution, including settings where
the integer budget is unchanged.  We retain the prompts, generation settings,
eight replicas, concurrency limit, frozen Bernoulli reward protocol, and
result-ready timing definition.  Each execution accepts exactly
\(N_j b^{\downarrow}_{j,b}\) rollouts.  The time increase in Table~\ref{tab:systems-benefit} is
\(\sum_j T_{\mathrm{U},j}(b^{\downarrow}_{j,b})/
\sum_j T_{\mathrm{U},j}(b)-1\), using the original Uniform measurements
in the denominator.

\paragraph{Rollout and token accounting.}
At the original budgets, every policy accepts exactly \(Nb\) rollouts.  Actual rollout counts add
completed-discarded and running-aborted requests to accepted rollouts.
The extra-rollout percentage in Table~\ref{tab:systems-compute} divides the
sum of HBN-async's actual counts minus Uniform's actual counts by the sum of
Uniform's actual counts within each budget.  Uniform's actual and accepted
counts coincide.  Each running-aborted request counts as one rollout.
For completed requests, we record prompt and completion token
counts.  For aborted requests, we record the observed prompt and generated
tokens before cancellation.  A queued cancellation with no observed model
compute contributes zero to the compute estimate.  Cancellation before vLLM
reports its first output can conceal already executed computation; hence
aborted compute and aggregate waste are observable lower bounds.

\paragraph{Sequence-length-aware FLOP proxy.}
Let \(P_r\) and \(C_r\) denote the prompt and observed completion lengths
of request \(r\), and write \(Q_r=P_r+C_r\).  We estimate its cost as
\begin{equation}
  \widehat F_r
  =2\theta_{\mathrm{core}}Q_r
   +2\theta_{\mathrm{head}}C_r
   +2a d_{\mathrm{attn}}Q_r(Q_r+1).
  \label{eq:systems-flop-proxy}
\end{equation}
Here a multiply--add counts as two FLOPs,
\(\theta_{\mathrm{core}}\) is the stored text-core parameter count excluding
input embeddings, the output head, vision, and multi-token-prediction
modules; \(\theta_{\mathrm{head}}\) is the output projection size (the
embedding matrix size for tied weights); \(a\) is the number of
full-attention layers; and \(d_{\mathrm{attn}}\) is query-head count times
head dimension.  Counts come from the checkpoint weights and
model configurations.  Table~\ref{tab:systems-flop-coefficients} lists the
coefficients.

\begin{table}[!htbp]
  \centering
  \small
  \begin{tabular}{@{}lrrrr@{}}
    \toprule
    Checkpoint & \(\theta_{\mathrm{core}}/10^9\)
      & \(\theta_{\mathrm{head}}/10^9\) & \(a\) & \(d_{\mathrm{attn}}\) \\
    \midrule
    Qwen2.5-3B  & 2.774774 & 0.311165 & 36 & 2048 \\
    Qwen2.5-7B  & 6.525622 & 0.544997 & 28 & 3584 \\
    Qwen2.5-32B & 31.206741 & 0.778568 & 64 & 5120 \\
    Qwen3.5-4B  & 3.570052 & 0.635699 &  8 & 4096 \\
    Qwen3.5-9B  & 6.919566 & 1.017119 &  8 & 4096 \\
    Qwen3.5-27B & 24.353202 & 1.271398 & 16 & 6144 \\
    \bottomrule
  \end{tabular}
  \caption{Coefficients for the estimated FLOP proxy.  Parameter counts are
  displayed in billions and rounded; attention width need not equal the
  model hidden size.}
  \label{tab:systems-flop-coefficients}
\end{table}

The first term approximates text-core work as parameter-proportional per
token, the second accounts for output projection, and the third approximates
causal query--key and attention--value operations over the sequence.
We omit explicit GDN recurrent-state updates, whose estimated contribution
to total FLOPs is marginal across the evaluated Qwen3.5 models and budgets.
This analytical proxy approximates model computation rather
than hardware execution cost.

We sum Eq.~\eqref{eq:systems-flop-proxy} over accepted requests to obtain
\(F^{\mathrm{useful}}\), over completed-discarded requests using their full
observed lengths, and over running-aborted requests using their recorded
partial lengths to obtain \(F^{\mathrm{waste}}\).  Useful and actual FLOP
increases in Table~\ref{tab:systems-compute} are ratios of summed costs at a
given budget.  The wasted-FLOP column divides HBN-async's wasted compute by
Uniform's useful FLOPs, which equal Uniform's actual FLOPs.  With this common
denominator, the useful-compute increase and wasted-compute contribution add
to the actual-compute increase.  In contrast, time is normalized separately within each
profile--budget setting as defined in Section~\ref{sec:system-overhead}, then
averaged with equal weight over the 107 profiles at each budget.  The
residual overhead is the ratio of mean times minus one, equivalently
\(\sum_j\widetilde T_{\text{HBN-async},j}/\sum_jT_{\mathrm{Uniform},j}-1\),
rather than the mean of per-profile ratios at that budget.  This keeps each
workload's model-specific reference time in the calculation.

\section{Additional Experimental Results}
\label{app:additional-results}

\subsection{Variance gains across profiles and failure boundary}
\label{app:profile-robustness}

We complement Section~\ref{sec:variance-reduction} with the question:
\emph{How broadly do HBN's gains hold across profiles, and where can it lose?}

\begin{figure}[!htbp]
  \centering
  \includegraphics[width=0.94\linewidth]{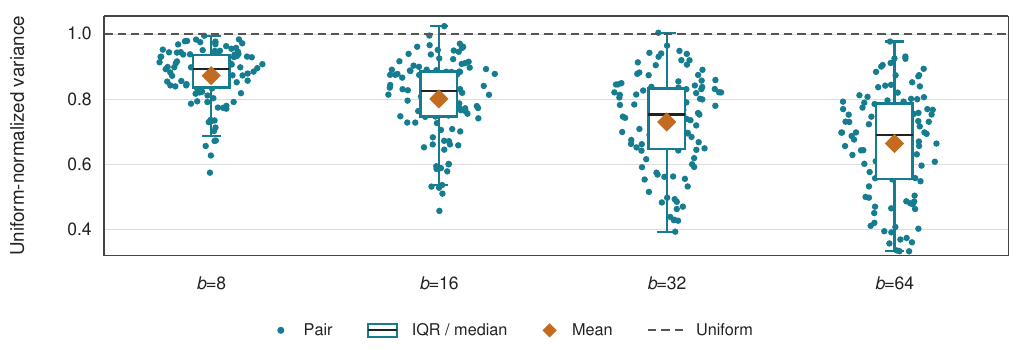}
  \caption{Distribution of HBN's Uniform-normalized variance across 107
  benchmark--checkpoint task-probability profiles.  Points are individual
  profiles, boxes show the interquartile range and median, diamonds show
  the mean, and the dashed line is Uniform.}
  \label{fig:hbn-pair-distribution}
\end{figure}

\paragraph{Gains are widespread, with a small adverse corner case.}
Figure~\ref{fig:hbn-pair-distribution} shows that the average gains are not
driven by a few profiles.  Requiring the Monte Carlo estimate plus
\(1.96\) standard errors to remain below one, HBN improves on Uniform for
107/107, 106/107, 106/107, and 107/107 profiles at
\(b=8,16,32,64\), respectively.

The only profile with an adverse result is MATH-100 with Qwen3.5-4B, where HBN
has mean variance ratios of \(1.024\) and \(1.004\) relative
to Uniform at \(b=16\) and \(32\), respectively.  This profile is highly
concentrated near the endpoints, with 96 of 100 tasks outside \([0.2,0.8]\).
Although its integer Oracle ratio of approximately \(0.774\) indicates
substantial allocative opportunity, a short pilot provides limited information
for reliably ranking the task variances.  HBN nevertheless reallocates
according to these noisy posterior differences, and the resulting
misallocation can hurt rather than improve performance.  Even in this worst
case, HBN is at most \(2.4\%\) worse than Uniform.  Thus, the observed
variance benefit is broad but not a per-profile guarantee: allocative
opportunity alone does not ensure that a limited pilot can exploit it.
Appendix~\ref{app:profile-case-study} provides a complementary detailed
AIME24--Qwen3.5-4B case study.

\subsection{Profile case study: AIME24 with Qwen3.5-4B}
\label{app:profile-case-study}

We examine AIME24 with Qwen3.5-4B as a heterogeneous
task-probability profile.  Among its 30 tasks, \(33.3\%\) have
\(\widehat p_i^{(K)}<0.2\), \(36.7\%\) have
\(\widehat p_i^{(K)}>0.8\), and four are empirical endpoints.

\begin{figure}[!htbp]
  \centering
  \setlength{\abovecaptionskip}{4pt}
  \setlength{\belowcaptionskip}{0pt}
  \includegraphics[width=\textwidth]{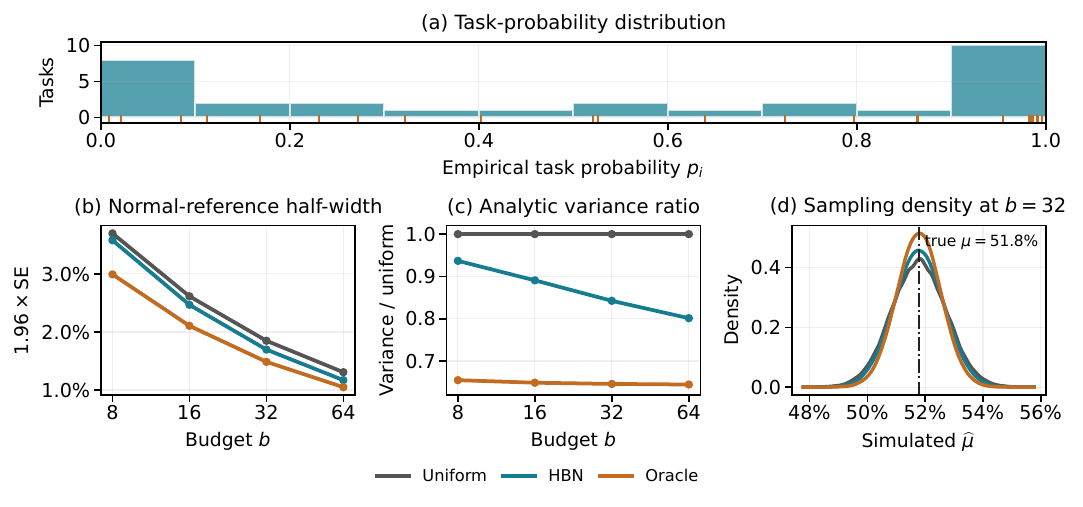}
  \caption{AIME24--Qwen3.5-4B under the jointly designed HBN policy.
  Panels show (a) the empirical task-probability distribution, (b) the
  normal-reference half-width, (c) the analytic fixed-profile variance ratio,
  and (d) sampling densities at \(b=32\) from 200,000 exact Bernoulli
  repetitions.  HBN uses the precommitted \((m,w)=(10,0.286)\) at \(b=32\);
  Oracle is the full-information integer Neyman allocation.}
  \label{fig:bayesian-neyman-case}
\end{figure}

For this profile, HBN's variance ratio falls from \(0.937\) at \(b=8\) to
\(0.801\) at \(b=64\), while the Oracle ratio is nearly flat at
\(0.655\)--\(0.644\).  At \(b=32\), HBN attains \(0.842\), compared with one
for Uniform and \(0.646\) for Oracle.  Figure~\ref{fig:bayesian-neyman-case}(d)
confirms that all three estimators remain centered, with HBN narrower than
Uniform under the fixed-weight estimator.

\subsection{A sequential posterior-sampling baseline}
\label{app:ts-neyman}

\paragraph{Motivation and adaptation.}
Inspired by TS-Neyman \citep{morikawa2026tsneyman}, we construct a sequential
posterior-sampling baseline for the same fixed-profile Bernoulli estimation
problem as HBN.  TS-Neyman allocates each new observation using a
posterior-sampled marginal variance-reduction index.  We retain this principle
while adapting the observation model and target weights to our setting;
we refer to the resulting baseline as Bernoulli TS-Neyman.

The original method uses a normal working model and an inverse-$\chi^2$
posterior over stratum variances.  Here outcomes are binary, so we instead
use independent $\mathrm{Beta}(\alpha,\alpha)$ working priors over task success
probabilities.  Beta--Bernoulli conjugacy provides posterior
samples of each task's sampling variance, $\widetilde p_i(1-\widetilde p_i)$.
Each task is an equally weighted stratum, with fresh independent Bernoulli
observations drawn from the fixed profile as in the main replay.

\paragraph{Sequential allocation and estimator.}
We first collect $m_0$ observations per task, charging all $Nm_0$ observations
to the total budget $Nb$.  After $t$ observations in total, let $n_i(t)$ and
$S_i(t)$ be the observation and success counts for task $i$.  At each step,
independently across tasks, we draw
$\widetilde p_i(t)\sim\mathrm{Beta}(\alpha+S_i(t),
\alpha+n_i(t)-S_i(t))$ and select
\begin{equation}
  i_{t+1}\in\arg\max_i
  \frac{\widetilde p_i(t)(1-\widetilde p_i(t))}
       {n_i(t)(n_i(t)+1)}.
  \label{eq:ts-bernoulli-index}
\end{equation}
The common target-weight factor $N^{-2}$ cancels from the index.
We break exact ties by task index, observe one new outcome for the selected
task, update its posterior, and repeat until $t=Nb$.
The final estimate is $\widehat\mu_{\mathrm{TS}}=
N^{-1}\sum_i S_i(Nb)/n_i(Nb)$, including the initial observations.
Because outcomes also influence subsequent
sample counts, this estimator need not be unbiased at finite budgets;
we therefore evaluate mean squared error (MSE), not variance alone.

\paragraph{Aligned replay and hindsight search.}
We use the same 107 fixed task-probability profiles and budgets
$b\in\{8,16,32,64\}$ as Section~\ref{sec:variance-reduction}.
For every profile and candidate, we simulate
8192 complete adaptive trajectories, recording the four budgets along each
trajectory where feasible.  HBN evaluates continuation variance
analytically, whereas TS simulates the interleaved allocation and observation
sequence.  We use common random numbers across TS candidates within each
profile.

We search $\alpha\in\{0.01,0.02,\ldots,0.20,0.25,0.50,1.00\}$,
matching the positive prior grid used for IBN, jointly with
$m_0\in\{2,5,10\}$, the warm-start values examined in the TS-Neyman paper.
The infeasible choice $m_0=10$ is excluded at $b=8$.
For each profile $j$, we normalize the empirical MSE by its analytic Uniform
MSE, $V_{\mathrm U,j}(b)=\sum_i p_{ji}(1-p_{ji})/(N_j^2b)$, and average
these ratios equally over the 107 profiles.  At each budget, the hindsight
envelope selects one common $(\alpha,m_0)$ minimizing this average.
HBN's reference variance equals its MSE
because its estimator is unbiased.  Selection and reporting use the same
replays, so the TS envelope may be optimistic due to Monte Carlo selection
noise.

\paragraph{Results.}
Table~\ref{tab:ts-neyman-envelope} reports the completed search over all 23
prior strengths and feasible warm starts.  HBN has lower mean
Uniform-normalized MSE at all four budgets, with gaps of
0.79--1.97 percentage points despite the hindsight selection afforded to TS.
The selected prior strengths are small ($\alpha=0.04$--$0.07$), favoring
endpoint-concentrated task probabilities.

This equal-rollout comparison does not charge TS's sequential
feedback overhead.
Each allocation decision in the evaluated TS policy depends on the preceding
observation, limiting concurrent generation when rewards become available only
after rollout completion.  HBN instead concentrates this dependence at the
pilot boundary, where speculative pre-allocation can overlap generation with
feedback collection.  Batched or asynchronous TS could relax the sequential
dependency, but would change the policy evaluated here.

\begin{table}[!htbp]
  \centering
  \begingroup
  \setlength{\tabcolsep}{0pt}
  \begin{tabular}{*{6}{>{\centering\arraybackslash}p{0.15\textwidth}}}
    \toprule
    $b$ & $\alpha$ & $m_0$ & TS-Neyman & HBN & $\Delta$ (pp) \\
    \midrule
    8  & 0.07 & 5  & 88.60\% & 87.25\% & 1.35 \\
    16 & 0.06 & 2  & 82.09\% & 80.12\% & 1.97 \\
    32 & 0.06 & 2  & 74.72\% & 73.03\% & 1.69 \\
    64 & 0.04 & 10 & 67.22\% & 66.44\% & 0.79 \\
    \bottomrule
  \end{tabular}
  \endgroup
  \caption{Bernoulli TS-Neyman hindsight envelope versus HBN.
  MSE entries are mean Uniform-normalized MSE across 107 profiles
  (lower is better).  Each budget selects one common $(\alpha,m_0)$;
  $\Delta$ is TS minus HBN in percentage points.}
  \label{tab:ts-neyman-envelope}
\end{table}

\section{Related Work}
\label{app:related-work}

\paragraph{Allocation and adaptive sampling.}
The oracle allocation $n_i\propto\sigma_i$ is Neyman allocation
\citep{neyman1934representative}, and inverse-probability estimation under
unequal sampling traces to \citet{horvitz1952generalization}.  Feasible
two-stage Neyman designs based on preliminary variance estimates are also
classical: \citet{sukhatme1975allocation} use fixed preliminary samples to
choose between proportional and estimated Neyman allocation, while
\citet{cai2024performance} show that allocation noise from a small fixed pilot
can make plug-in Neyman allocation less efficient than balanced allocation.

Many recent unknown-variance methods instead update allocation online.  Modern
active measurement constructs unbiased per-round estimates
\citep{hamilton2025active}, while TS-Neyman uses posterior-sampled marginal
variance reductions for sequential stratified model evaluation and reports
severe sparse-pilot failures of plug-in rules \citep{morikawa2026tsneyman}.
Several close analyses emphasize eventual large-budget behavior:
\citet{cai2024performance} hold the pilot fixed while the main wave tends to
infinity, TS-Neyman proves convergence to the Neyman target and asymptotic
optimality as its sequential budget grows, and EST-IVWE matches its oracle rate
up to lower-order terms in the total budget \citep{lee2026instanceoptimal}.
These results complement our operational target: \(b=8\)--\(64\) target-model
rollouts per task, where short pilots provide limited task-level evidence.
Online policies can adapt as feedback arrives, but their parallel execution
must account for delayed feedback and in-flight requests.  Our two-stage
design fixes the continuation allocation after one complete pilot, leaving
execution before that barrier to speculation.

\paragraph{Repeated stochastic LLM evaluation.}
ReliableEval treats prompt-preserving perturbations as a source of evaluation
randomness and uses method-of-moments analysis to determine how many repeats
are needed for stable conclusions \citep{lior2025reliableeval}.  More directly,
\citet{saha2026llmjudge} and \citet{lee2026instanceoptimal}
study allocation for
vectors of LLM-as-a-judge scores, whereas we estimate one fixed-benchmark mean
from target-model rollouts.  Saha et al. minimize worst-coordinate
($\ell_\infty$) error for a single judge, yielding the oracle allocation
$n_i\propto\sigma_i^2$.  Their unknown-variance ROBIN-HOOD method follows a
uniform warm-up with round-by-round allocation using upper confidence bounds
on the variances.  Lee et al. extend the vector problem to multiple judges with
heterogeneous costs and general $\ell_p$ error; with one equal-cost judge and
$p=2$, their oracle reduces to Neyman allocation, $n_i\propto\sigma_i$.
Our setting instead targets a fixed-benchmark mean from
low-budget Bernoulli rollouts, with hierarchical pooling of pilot evidence
across tasks.

\paragraph{Active and proxy-assisted model evaluation.}
Active Testing and Active Surrogate Estimators reduce label cost by selecting
which examples to label and correcting the resulting adaptive sample
\citep{kossen2021active,kossen2022surrogate}.  Factorized Active Querying uses
historical model--question factors to choose a subset of questions and an
augmented inverse-probability estimator to recover fixed-bank accuracy with
statistical guarantees \citep{wu2026faq}.  Approximate Neyman Allocation
partitions a finite input pool using surrogate semantic entropy and allocates
target labels across strata according to proxy Neyman weights
\citep{liu2026approximate}; the surrogate generations are outside the reported
target-label budget, and there is no paid target pilot to recycle.  MultiPPI
instead allocates budget across expensive measurements and multiple cheaper,
correlated autoraters or proxies \citep{cowenbreen2026multipppi}.
Benchmark-compression methods such as tinyBenchmarks and Efficient Benchmarking
reduce the evaluated suite itself or use coarse-to-fine evaluation
\citep{maiapolo2024tinybenchmarks,perlitz2024efficient}.  These approaches
change which items or measurement sources are queried.  We retain every task
in the fixed estimand and change only its number of target-model replications.

\paragraph{Adaptive inference and training compute.}
A neighboring line of work treats repeated generations as compute used to
improve the model's output rather than to estimate its performance.  Strategic
Scaling formulates test-time allocation across queries as a bandit problem and
maximizes the fraction of queries ultimately answered correctly under a fixed
generation budget \citep{zuo2026strategic}.  On the training side, VIGOR
progressively assigns RLVR rollouts to examples with high group-reward variance
because that variance controls the training signal \citep{jiang2026vigor}.
These methods and ours exploit heterogeneous returns to additional rollouts,
but their objectives are answer quality or learning progress; ours is the
precision of an unbiased fixed-benchmark mean.

\end{document}